\documentclass[letterpaper, 10 pt, conference]{ieeeconf}  

\IEEEoverridecommandlockouts                              

\usepackage{graphics} 
\usepackage{epsfig} 
\usepackage{amsmath} 
\usepackage{amssymb}  
\DeclareMathOperator*{\argmin}{arg\,min}
\usepackage{algorithm}
\usepackage{algorithmic}
\usepackage[font=small,labelfont=bf]{caption}
\usepackage{hyperref}
\title{
Manipulating Highly Deformable Materials Using a Visual Feedback Dictionary
}

\author{Biao Jia, Zhe Hu, Jia Pan, Dinesh Manocha \\
\url{http://gamma.cs.unc.edu/ClothM/}
\thanks{Biao Jia and Dinesh Manocha are with the Department of Computer Science, University of North Carolina at Chapel Hill. E-mail: {\tt\small \{biao, dm\}@cs.unc.edu}.}
\thanks{Zhe Hu and Jia Pan are with the Department of Mechanical Biomedical Engineering, City University of Hong Kong.}
}

\begin{document}
\maketitle

\begin{abstract}
The complex physical properties of highly deformable materials such as clothes pose significant challenges fanipulation systems. We present a novel visual feedback dictionary-based  method for manipulating defoor autonomous robotic mrmable objects towards a desired configuration. Our approach is based on visual servoing and we use an efficient technique to extract key features from the RGB sensor stream in the form of a histogram of deformable model features. These histogram features serve as high-level representations of the state of the deformable material. Next, we collect manipulation data and use a visual feedback dictionary that maps the velocity in the high-dimensional feature space to the velocity of the robotic end-effectors for manipulation. We have evaluated our approach on a set of complex manipulation tasks and human-robot manipulation tasks on different cloth pieces with varying material characteristics. 
\end{abstract}

\section{Introduction}
\label{intro}
The problem of manipulating highly deformable materials such as clothes and fabrics frequently arises in different applications. These include laundry folding~\cite{Miller:2011:AGA}, robot-assisted dressing or household chores~\cite{Kapusta:2016:DDH, Gao:2016:IPO}, ironing~\cite{Li:2016:MSS}, coat checking~\cite{Twardon:2015:ISC},  sewing~\cite{Schrimpf:2012:RSI}, and transporting large materials like cloth, leather, and composite materials~\cite{kruse2015collaborative}. Robot manipulation has been extensively studied for decades and there is extensive work on the manipulation of rigid and deformable objects.  Compared to the manipulation of a rigid object, the state of which can be completely described by a six-dimensional configuration space, the manipulation of a deformable object is more challenging due to its very high configuration space dimensionality. The resulting manipulation algorithm needs to handle this dimensional complexity and maintain the tension to perform the task.

One practical approach to dealing with general deformable object manipulation is based on visual servoing,~\cite{Alarcon:2013:MFV, Alarcon:2016:ADM}. At a broad level, these servoing techniques use perception data captured using cameras and formulate a  control policy mapping to  compute the velocities of the robotic end-effectors in real-time. 
However, a key issue in these methods is to automatically extract key low-dimensional features of the deformable material that can be used to compute a mapping. The simplest methods use manual or other techniques to extract features corresponding to line segments or curvature from a set of points on the surface of the deformable material. 
In addition, we need appropriate mapping algorithms based on appropriate low-dimensional features. 
Current methods may not work well while performing complex tasks or when the model undergoes large deformations. 
Furthermore, in many human-robot systems, the deformable material may undergo unknown perturbations and it is important to design robust manipulation strategies~\cite{Gao:2016:IPO,Kruse:2015:CHR}.


\begin{figure}[t]
  \centering
  \includegraphics[width=0.5\textwidth]{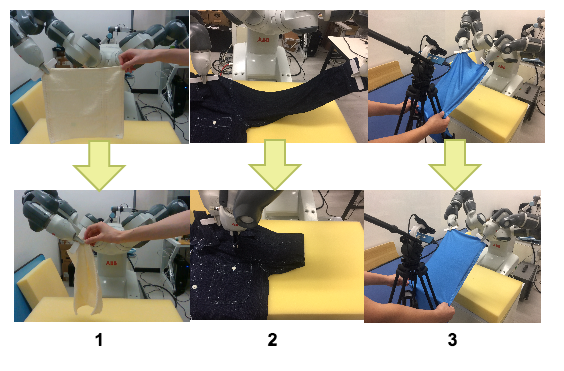}
    \vspace*{-0.12in}
  \caption{
  {\bf Manipulation Benchmarks:} We highlight the real-time performance of our algorithm on three tasks: (1) human-robot jointly folding a cloth with one hand each; (2) robot folding a cloth with two hands; (3) human-robot stretching a cloth with four combined hands. The top row shows the initial state for each task and the bottom row is the final state. Our approach can handle the perturbations due to human movements.
  }
  \label{fig:benchmark1}
  \vspace*{-0.12in}
\end{figure}

\noindent{\bf Main Results:}
In this paper, we present a novel feature representation, a histogram of oriented wrinkles (HOW), to describe the shape variation of a highly deformable object like clothing. These features are computed by applying Gabor filters and extracting the high-frequency and low-frequency components.
We precompute a visual feedback dictionary using an offline training phase that stores a mapping between these visual features and the velocity of the end-effector. At runtime, we automatically compute the goal configurations based on the manipulation task and use sparse linear representation to compute the velocity of the controller from the dictionary (Section III).  Furthermore, we extend our approach so that it can be used in human-robot collaborative settings. Compared to prior manipulation algorithms, the novel components of our work include:
\begin{itemize}
\item A novel histogram feature representation of highly deformable materials (HOW-features) that are computed directly from the streaming RGB data using Gabor filters (Section ~\ref{sec:feat}).
\item A sparse representation framework using a visual feedback dictionary, which directly correlates the histogram features to the control instructions (Section ~\ref{sec:learning}).
\item The combination of deformable feature representation, a visual feedback dictionary, and sparse linear representations that enable us to perform complex manipulation tasks, including human-robot collaboration, without significant training data (Section ~\ref{sec:goal}). 
\end{itemize}
We have integrated our approach with an ABB YuMi dual-arm robot and a camera for image capture and used it to manipulate different cloth materials for different tasks. We highlight the real-time performance for tasks related to stretching, folding, and placement (Section \ref{sec:exp}).


\section{Related Work}
\label{sec:related}
Many techniques have been proposed for the automated manipulation of flexible materials. 
Some of them have been designed for specific tasks, such as peg-in-hole and laying down tasks with small elastic parts~\cite{bodenhagen2014adaptable} or wrapping a cloth around a rigid surface~\cite{berenson2013manipulation}. There is extensive work on folding laundry using pre-planned materials.
In this section, we give a brief review of prior work on deformable object representation, manipulation, and visual servoing.

\subsection{Deformable Object Representation}
The recognition and detection of deformable object characteristics is essential for manipulation tasks.  There is extensive work in computer vision and related areas on tracking features of deformable models. Some of the early work is based on active deformable models~\cite{sullivan1996using}. Ramisa \emph{et al.}~\cite{ramisa2012using} identify the grasping positions on a cloth with many wrinkles using a bag-of-features-based detector. Li \emph{et al.}~\cite{li2014recognition} encode the category and pose of a deformable object by collecting a large set of training data in the form of depth images from different view points using offline simulation.  

\subsection{Deformable Object Manipulation}
Robotic manipulation of general deformable objects relies on a combination of different sensor measurements. The RGB images or RGB-Depth data are widely used for deformable object manipulation~\cite{Miller:2011:AGA,Li:2016:MSS,Alarcon:2016:ADM}. Fiducial markers can also be printed on the deformable object to improve the manipulation performance~\cite{bersch2011bimanual}.  In addition to visual perception, information from other sensors can also be helpful, like the use of contact force sensing to maintain the tension~\cite{kruse2015collaborative}.
In many cases, simulation techniques are used for manipulation planning. 
Clegg \emph{et al.}~\cite{clegg2017learning} use reinforcement learning to train a controller for haptics-based manipulation. 
Bai \emph{et al.}~\cite{bai2016dexterous} use physically-based optimization to learn a deformable object manipulation policy for a dexterous gripper. 
McConachie \emph{et al.}  ~\cite{mcconachie2017bandit} formulate model selection for deformable object manipulation and introduces a utility metric to measure the performance of a specific model.
These simulation-based approaches need accurate deformation material properties, which can be difficult to achieve.
Data-driven techniques have  been used to design the control policies for deformable object manipulation. 
Yang \emph{et al.}~\cite{yang2017repeatable} propose an end-to-end framework to learn a control policy using deep learning techniques for folding clothes.
Cusumano-Towner \emph{et al.} ~\cite{cusumano2011bringing} learn a Hidden Markov Model (HMM) using a sequence of manipulation actions and observations. 

\subsection{Visual Servoing for Deformable Objects}
Visual servoing techniques ~\cite{chaumette2006visual,hutchinson1996tutorial} aim at controlling a dynamic system using visual features extracted from images. 
They have been widely used in robotic tasks like manipulation and navigation. 
Recent work includes the use of histogram features for rigid objects ~\cite{bateux2017histograms}.
Sullivan \emph{et al.}~\cite{sullivan1996using} use a visual servoing technique to solve the deformable object manipulation problem using active models. 
Navarro-Alarcon \emph{et al.}~\cite{Alarcon:2013:MFV,Alarcon:2016:ADM} use an adaptive and model-free linear controller to servo-control soft objects, where the object's deformation is modeled using a spring model~\cite{Hirai:2000:ISP}. 
Langsfeld \emph{et al.}~\cite{Langsfeld:2016:OLP} perform online learning of part deformation models for robot cleaning of compliant objects. Our goal is to extend these visual servoing methods to perform complex tasks on highly deformable materials. 


\begin{table}
\begin{tabular}{p{.1\textwidth}p{.85\textwidth}}
$m$ & state of the deformable object \\
$r$ & robot's end-effector configuration \\
$v$ & robot's end-effector velocity, $v=\dot{r}$ \\
$I(w,h)$ & image from the camera, of size $(w_I,h_I)$\\ 
$s(I)$ & HOW-feature vector extracted from image $I$ \\
$d_i(I)$ & the $i$-{th} deformation kernel filter \\


$I^{(t)},r^{(t)}$ & time index t of images and robot configurations \\
$\{\Delta s^{(i)}\}$,$\{\Delta r^{(i)}\}$ &  features and labels of the visual feedback dictionary \\
$\rho(\cdot, \cdot)$  & error function between two items \\
$\lambda$ & positive feedback gain \\ 

$L$ & interaction matrix linking velocities of the feature   \\
& space to the end-effector configuration space \\
$F$ & interaction function linking velocities of  the feature \\
& space to the end-effector configuration space \\
$p_j(i)$ & $j_{th}$ histogram of value $i$ \\
$N_{dof,I,r,d}$ & number of degrees of freedom of the manipulator\\ 
	& images $\{I^{(t)}\}$, samples $\{r^{(t)}\}$ ,filters $\{d^{(t)}\}$ \\
$C_{fr}$ & constant of frame rate \\
$r^*, s^*, m^*$ & desired target configuration, feature, state\\
$\hat{r}, \hat{s}, \hat{m}$ & approximated current configuration, feature, state\\
\end{tabular}
\caption{Symbols used in the paper}
\label{tab:nomenclature}
\vspace*{-0.12in}
\end{table}

\section{Overview}
\label{sec:overview}
In this section, we introduce the notation used in the paper. We also present a formulation of the deformable object manipulation problem. Next, we give a brief background 
on visual servoing and the feedback dictionary. Finally, we give an overview of our deformable manipulation algorithm that uses a visual feedback dictionary. 

\subsection{Problem Formulation}
The goal of manipulating a highly deformable material is to drive a soft object towards a given target state ($m^*$) from its current state ($m$). The state of a deformation object ($m$)  can be complex due to its high dimensional configuration. In our formulation, we do not represent this state explicitly and treat it as a hidden variable.  Instead, we keep track of the deformable object's current state in the feature space ($s$) and its desired state ($s^*$), based on HOW-features.

The end-effector's configuration $r$ is represented using the Cartesian coordinates and the orientations of end-effectors or the degree of each joint of the robot. When $r$ corresponds to the Cartesian coordinates of the end-effectors, an extra step is performed by the inverse kinematics motion planner \cite{beeson2015trac} to map the velocity $v$ to the actual controlling parameters.

The visual servo-control is performed by computing an appropriate velocity ($v$) of the end-effectors. Given the visual feedback about the deformable object,
the control velocity ($v$) reduces the error in the feature space ($s-s^*$). 
After continuously applying feedback control, the robot will manipulate the deformable object toward its goal configuration. In this way, the feedback controller can be formulated as computing a mapping between the velocity in the feature space of the deformable object and the velocity in the end-effector's configuration space ($r$):

\begin{equation}
r^* - r = -\lambda F (s - s^*)
\label{eq:interaction}
\end{equation}
where $F$ is an interaction function that is used to map the two velocities in different spaces and $\lambda$ is the feedback gain. 
This formulation works well only if some standard assumptions related to the characteristics of the deformable object hold. These include infinite flexibility with no energy contribution from bending, no dynamics, and being subject to gravity and downward tendency (see details in ~\cite{van2010gravity}). For our tasks of manipulating highly deformable materials like clothes or laundry  at a low speed, such assumptions are valid.


\begin{figure}[t]
  \centering
  \includegraphics[width=0.4\textwidth]{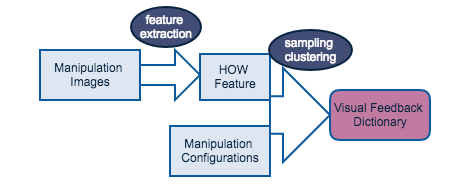}
  \caption{ 
  {\bf Computing the Visual Feedback Dictionary:}
  {The input to this offline process is the recorded manipulation data with images and the corresponding configuration of the robot's end-effector. The output is a visual feedback dictionary, which links the velocity of the features and the controller.}
  }
  \label{fig:offline}
\end{figure}

\subsection{Visual Servoing}
\label{subsec:background}
In this section we give a brief overview of visual servoing~\cite{chaumette2006visual,collewet2011photometric,bateux2017histograms, bateux2015direct}, which is used in our algorithm.
In general-purpose visual servoing, the robot wants to move an object from its original configuration ($r$) towards a desired configuration ($r^*$). In our case, the object's configuration ($r$) also corresponds to the configuration of the robot's end-effector, because the rigid object is fixed relative to the end-effectors during the manipulation task. These methods use a cost function $\rho(\cdot, \cdot)$ as the error measurement in the image space and the visual servoing problem can be formulated as an optimization problem:
\begin{equation}
\hat{r^*} = \argmin_r \rho(r, r^*), \label{eq:optimization}
\end{equation}
where $\hat{r^*}$ is the configuration of the end-effector after the optimization step and is the closest possible configuration to the desired position ($r^*$). In the optimal scenario, $\hat{r^*} = r^*$. 

Let $s$ be a set of HOW-features extracted from the image. Depending on the configuration $r$, the visual servoing task is equivalent to minimizing the Euclidean distance in the feature space and this can be expressed as:
\begin{equation}
\hat{r^*} = \argmin_r ((s(r)-s^*)^T (s(r)-s^*)),
\end{equation}
where $s^* = s(r^*)$ is the HOW-feature corresponding to the goal configuration ($r^*$). 
The visual servoing problem can be solved by iteratively updating the velocity of the end-effectors according to the visual feedback:
\begin{equation}
\label{eq:velocity}
v = -\lambda L^+_s (s  - s^*),
\end{equation}
where $L^+_s$ is the pseudo-inverse of the interaction matrix $L_s= \frac{\partial s}{\partial r}$. It corresponds to an interaction matrix  that links the variation of the velocity in the feature space $\dot{s}$ to the velocity in the end-effector's configuration space. 
$L^+_s$ can be computed offline by training data defined in \cite{bateux2017histograms}.

\subsection{Visual Feedback Dictionary}
The visual feedback dictionary corresponds to a set of vectors with instances of visual feedback $\{ \Delta s^{(i)} \}$ and  the corresponding velocities of the end-effectors $\{ \Delta r^{(i)} \}$, where $\{ \Delta s^{(i)} \}= (s - s^*)$. Furthermore, we refer to each instance $\{\Delta s^{(i)},  \Delta r^{(i)}\}$ as the {\em  visual feedback word}.   This dictionary is computed from 
 the recorded manipulation data. The input includes a set of images ($\{I^{(t)}\}$) and the end-effector configurations ($\{r^{(t)}\}$). Its output is computed as ($\{\{ \Delta s^{(i)} \}, \{ \Delta r^{(i)} \}\}$).
We compute this dictionary during an offline learning phase using sampling and clustering methods (see Algorithm 2 for details), and use this dictionary at runtime to compute the velocity of the controller by computing the sparse decomposition of a feedback $\Delta s$.
More details about computing the visual feedback dictionary are give in Algorithm 2.



\begin{figure}[t]
  \centering
  \includegraphics[width=0.4\textwidth]{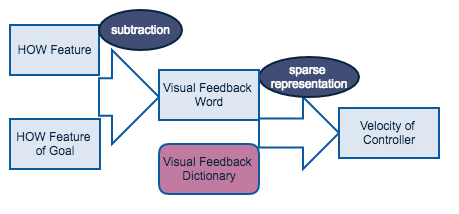}
  \caption{
  {\bf Runtime Algorithm:} 
  {The runtime computation consists of two stages.
  We extract the deformable features (HOW-features) from the visual input and computes the visual feedback word by subtracting the extracted  features from the features of the goal configuration. We  apply the sparse representation and compute the velocity of the controller for manipulation.}
  }
  \label{fig:online}
\end{figure}

\subsection{Offline and Runtime Computations}
Our approach computes the visual dictionary using an offline process (see Fig. \ref{fig:offline}). Our runtime algorithm consists of two components. Given the camera stream, we extract the HOW-features from the image ($s(I)$) and compute the corresponding velocity of the end-effector using an appropriate mapping.
For the controlling procedure, we use the sparse representation method  to compute the interaction matrix, as opposed to directly solving the   optimization problem  (Equation \ref{eq:optimization}). In practice, the sparse representation method is more efficient.
The runtime phase performs the actual visual servoing with the current image at time $t$ ($I^{(t)}$), the visual feedback dictionary ($\{\{ \Delta s^{(i)} \}, \{ \Delta r^{(i)} \}\}$), and the desired goal configurations given by ($I^*$) as  the input. 


\section{Histogram of Deformation Model Feature}
\label{sec:feat}
\begin{figure*}[!htbp]
  \centering
  \includegraphics[width=0.9\textwidth]{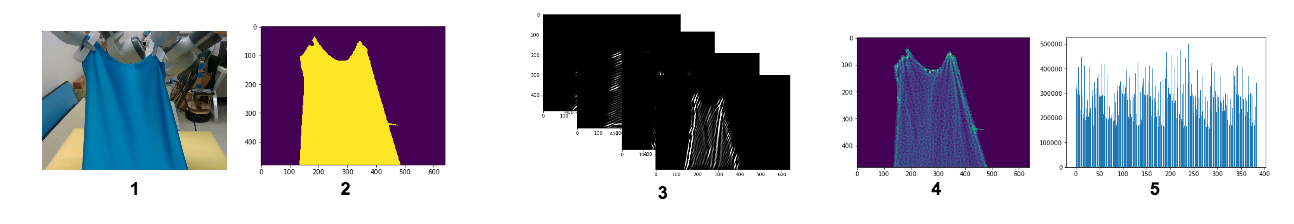}
  \caption{
  {\bf Pipeline for HOW-feature Computation:}
  {We use the following stages for the input image (1):
  (2) Foreground segmentation using Gaussian mixture;
  (3) Image filtering with multiple orientations and wavelengths of Gabor Kernel;
  (4) Discretization of the filtered images to form grids of histogram;
  (5) Stacking the feature matrix to a single column vector.
  }
  }
  \label{fig:feature}
\end{figure*}

In this section, we present our algorithm to compute the HOW-features from the camera stream. These are low-dimensional features of the highly deformable material.   
The pipeline of our HOW-feature computation process is shown in Figure \ref{fig:feature}. Next, we explain each of these stages in detail.

\subsection{Foreground Segmentation}
To find the foreground partition of a cloth, we apply the Gaussian mixture  algorithm~\cite{lee2005effective} on the RGB data captured by the camera. The intermediate result of segmentation is shown in Figure \ref{fig:feature}(2). 

\subsection{Deformation Enhancement}
To model the high dimensional characteristics of the highly deformable material, we use deformation enhancement. This is based on the perceptual observation that most deformations can be modeled by shadows and shape variations.
Therefore, we extract the features corresponding to shadow variations  by applying a Gabor transform~\cite{lee1996image} to the RGB image. This results in the enhancement of the ridges, wrinkles, and edges (see Figure \ref{fig:feature}). 
We convolve the $N$ deformation filters $\{d_i\}$  to the image $I$ and represent the result as $\{d_i(I)\}$.

In the spatial domain, a 2D Gabor filter is a Gaussian kernel function modulated by a sinusoidal plane wave~\cite{daugman1985uncertainty} and it has been used to detect wrinkles~\cite{yamazaki2009cloth}.  The 2D Gabor filter 
can be represented as follows:
\begin{equation}
g(x,y;\lambda, \theta, \phi, \sigma, \gamma) = \exp(-\frac{ {x'}^2 + \gamma ^2 {y'}^2}{2 \sigma ^2}) \sin(2 \pi \frac{x'}{\lambda} + \phi),
\end{equation}
where $x' = x \cos\theta + y \sin\theta$, $y' = -x \sin\theta + y\cos\theta$, $\theta$ is the orientation of the normal to the parallel stripes of the Gabor filter, $\lambda$ is the wavelength of the sinusoidal factor, $\phi$ is the phase offset, $\sigma$ is the standard deviation of the Gaussian, and $\gamma$ is the spatial aspect ratio. When we apply the Gabor filter to our deformation model image, the choice of  wavelength ($\lambda$) and orientation ($\theta$) are the key parameters with respect to the wrinkles of deformable materials. As a result, the deformation model features consist of multiple Gabor filters ($d_{1 \cdots n}(I)$) with different values of wavelengths ($\lambda$) and orientations ($\theta$).

\subsection{Grids of Histogram}
A histogram-based feature is an approximation of the image which can reduce the data redundancy and extract a high-level representation that is robust to local variations in an image. Histogram-based features have been adopted to achieve a general framework for photometric visual servoing~\cite{bateux2017histograms}.
Although the distribution of the pixel value can be represented by a histogram, it is also significant to represent the position in the feature space of the deformation to achieve the manipulation task. Our approach is inspired by the study of grids of Histogram of Oriented Gradient~\cite{dalal2005histograms}, which is computed on a dense grid of uniformly spatial cells. 

We compute the grids of histogram of deformation model feature by dividing the image into small spatial regions and accumulating local histogram of different filters ($d_i$) of the region. For each grid, we compute the histogram in the region and represent it as a matrix. We vary the grid size and compute matrix features for each size. Finally, we represent the entries of a matrix as a column feature vector. 
The complete feature extraction process is described in Algorithm \ref{alg:feat}.

\begin{algorithm}[!htp!]
  \caption{Computing HOW-Features}
  \label{alg:feat}
  \begin{algorithmic}[1]
    \REQUIRE image $I$ of size $(w_I, h_I)$, deformation filtering or Gabor kernels $\{d_1 \cdots d_{N_d}\}$, grid size set$\{g_1,\cdots,g_{N_g}\}$.
    \ENSURE feature vector $s$
	\FOR {$i=1,\cdots,N_d$}
    \FOR {$j=1,\cdots,N_g$}
      \FOR {$(w,h)=(1,1),\cdots, (w_I,h_I)$}
          \STATE $(x,y) =(\textsc{trunc}(w / g_j), \textsc{trunc}(h / g_j))$ //  compute the indices using truncation
          \STATE $s_{i,j,x,y} = s_{i,j,x,y} + d_i(I(w,h))$ //add the filtered pixel value to the specific bin of the grid
	  \ENDFOR
    \ENDFOR
    \ENDFOR
    \RETURN $s$
  \end{algorithmic}
\end{algorithm}

The HOW-feature has several advantages. It captures the deformation structure, which is based on the characteristics of the local shape.
Moreover, it uses a local representation that is invariant to local geometric and photometric transformations. This is useful when the translations or rotations are much smaller than the local spatial or orientation grid size.

\section{Manipulation using Visual Feedback Dictionary}
\label{sec:learning}
In this section, we present our algorithm for computing the visual feedback dictionary. At runtime, this dictionary is used to compute the corresponding velocity ($\Delta r$) of the controller based on the visual feedback ($\Delta s(I)$). 

\begin{figure}[b]
  \centering
  \includegraphics[width=0.45\textwidth]{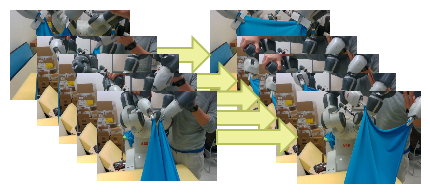}
    \vspace*{-0.12in}
  \caption{
  {\bf Visual Feedback Dictionary:} 
  {The visual feedback word 
  is defined by the difference between the visual features $\Delta s = s - s^*$ and the controller positions $\Delta r = r - r^*$. The visual feedback dictionary $\{\{\Delta s^{(i)}\},\{\Delta r^{(i)}\}\}$ consists of visual feedback words computed. We show the initial and final states on the left and right, respectively. }
  }
  \label{fig:dictionary}
  \vspace*{-0.12in}
\end{figure}

\subsection{Building Visual Feedback Dictionary}
As shown in Figure \ref{fig:offline}, the inputs of the offline training phase are a set of images and end-effector configurations ($\{I^{(t)}\}$, $\{r^{(t)}\}$) and the output is the visual feedback dictionary ($\{\{\Delta s^{(i)}\},\{ \Delta r^{(i)}\}\}$). 

For the training process, the end-effector configurations, ($\{r^{(t)}\}$), are either collected by human tele-operation or generated randomly. A single configuration ($r^{(i)}$) of the robot is a column vector of length $N_{dof}$, the number of degrees-of-freedom to be controlled. $r \in \mathbb{R}^{N_{dof}}$ and its value is represented in the configuration space.

In order to compute the mapping ($F_H$) from the visual feedback to the velocity, we need to transform the configurations $\{r^{(t)}\}$ and image stream $\{I^{(t)}\}$ into velocities $\{\Delta r^{(t)}\}$ and  the visual feature feedback $\{\Delta s^{(t)}\}$. One solution is to select a fixed time step $\Delta t$ and to represent the velocity in both the feature and the configuration space as:
{\small
\[
\Delta r^{(t)} = \frac{r^{(t+\frac{\Delta t}{2})} - r^{(t-\frac{\Delta t}{2})}}{\Delta t / C_{fr}}; \ 
\Delta s(I^{(t)}) = \frac{ s(I^{(t+\frac{\Delta t}{2})}) - s(I^{(i-\frac{\Delta t}{2})})}{\Delta t / C_{fr}}
\]
}
\noindent where $C_{fr}$ is the frame rate of the captured video.

However, sampling by a fixed time step ($\Delta t$) leads to a limited number of samples ($N_{I} - \Delta t$) and can result in over-fitting. To overcome this issue, we break the sequential order of the time index to generate more training data from  ${I^{(t)}}$ and ${r^{(t)}}$. 
In particular, we assume the manipulation task can be observed as a Markov process~\cite{baum1972inequality} and each step is independent from every other. In this case, the sampling rates are given as follows, (when the total sampling amount is $n$):
{\small
\[
\Delta r^{(t)} = \frac{r^{(p_{t})}- r^{(p_{t+n})}} { (p_{t} - p_{t+n}) / C_{fr}}; \ 
\Delta s(I^{(t)}) = \frac{s(I^{(p_{t})}) - s(I^{(p_{t+n})})} { (p_{t} - p_{t+n}) / C_{fr}}
\]
}
\noindent where $p_{1,\cdots, 2n}$ is a set of indices randomly generated, and $p_t \in [1 \cdots N_I]$.
In order to build a more concise dictionary, we also apply K-Means Clustering~\cite{hartigan1979algorithm} on the feature space, which enhance the performance and prevent the over-fitting problem.   

In practice, the visual feedback dictionary can be regarded as an approximation of the interaction function $F$ (see Equation \ref{eq:interaction}). The overall algorithm to compute the dictionary is given in Algorithm \ref{alg:training}.

\subsection{Sparse Representation}
At runtime, we use sparse linear representation~\cite{donoho2003optimally} to compute the velocity of the controller from the visual feedback dictionary. 
These representations tend to assign zero weights to most irrelevant or redundant features and  are used to find a small subset of most predictive features in the high dimensional feature space.
Given a noisy observation of a feature at runtime ($s$) and the visual feedback dictionary constructed by features $\{\Delta s^{(i)}\}$ with labels $\{\Delta r^{(i)}\}$, we represent $\Delta s$ by $\hat{\Delta s}$, which is a sparse linear combination of $\{\Delta s^{(i)}\}$, where $\beta$ is the sparsity-inducing $L_1$ term. To deal with noisy data, we rather use the $L_2$ norm on the data-fitting term and formulate the resulting sparse representation as:
\begin{equation}
\hat{\beta} = \argmin_\beta( ||min(\Delta s - \sum_i \beta_i \Delta s^{(i)} )||_2^2+\alpha ||\beta||_1),  \label{alpha1}
\end{equation}
where $\alpha$ is a slack variable that is used to balance the trade-off between fitting the data perfectly and using a sparse solution.
The sparse coefficient $\beta^*$ is computed using a minimization formulation:
\begin{equation}
   \beta^* =  \argmin_\beta (\sum_{i} {\rho(\Delta s_i^* - \sum_{j} {\beta_j \Delta s^{(j)}_i })} \label{alpha2}
        + \alpha \sum_{j} {||\beta_j||_1}) 
\end{equation}
After $\hat{\beta}$ is computed, the observation $\hat{\Delta s}$ and the probable label $\hat{\Delta r}$ can be reconstructed by the visual feedback dictionary :
\begin{eqnarray}
\hat{\Delta s} = \sum_i \hat{\beta_i} \Delta s^{(i)} \quad \quad 
\hat{\Delta r} = \sum_i \hat{\beta_i} \Delta r^{(i)}
\end{eqnarray}
The corresponding  ${\Delta r}^*$ of the $i-$th  DOF in the configuration is given as: 
\begin{equation}
   \Delta r_i^* = \sum_{j} \beta^*_j \Delta r_i^{(j)}, 
\end{equation}
where $\Delta s^{(j)}_i$ denotes the $i-$th datum of the $j-$th feature , $\Delta s^*_i$ denotes the value of the response, and the norm-1 regularizer $ \sum_{j} {||\beta_j||_1}$  typically results in a sparse solution in the feature space. 

\begin{algorithm}[t]
  \caption{Building the Visual Feedback Dictionary}
  \label{alg:training}
  \begin{algorithmic}[1]
    \REQUIRE image stream $\{I^{(t)}\}$ and positions of end-effectors $\{r^{(t)}\}$ with sampling amount $n$, dictionary size $N_{dic}$
    \ENSURE Visual Feedback Dictionary $\{\{\Delta s_d^{(i)}\},\{ \Delta r_d^{(i)}\}\}$
    \STATE $\{\Delta s_d\} = \{\}$, $\{\Delta r_d\} = \{\}$
    \STATE $p = N_I \textsc{rand}( 2n ) $ // generate random indices for sampling
    \FOR {$i = 1, \cdots, n$}
    	\STATE $\Delta s^{(i)}$ = $s(I^{(p(i))}) - s(I^{(p(i+n))})$ // sampling
    	\STATE $\Delta r^{(i)}$ = $r^{(p(i))} - r^{(p(i+n))}$ // sampling
    \ENDFOR
    \STATE  $centers = \textsc{K-MEANS}(\{\Delta s^{(i)}\}, N_{dic})$ // compute the centers of the feature set for clustering
    \FOR {$i = 1, \cdots, N_{dic}$}
    	\STATE $j = \argmin_i (centers[i] - s^{(i)})$
        \STATE $\{\Delta s_d\} = \{\Delta s_d, \Delta s^{(j)}\}$ $\{\Delta r_d\} = \{\Delta r_d, \Delta r^{(j)}\}$
    \ENDFOR
   \RETURN $\{\Delta s_d\},\{\Delta r_d\}$ 
  \end{algorithmic}
\end{algorithm}


\subsection{Goal Configuration and Mapping}
\label{sec:goal}
We compute the goal configuration and the corresponding HOW-features based on the underlying manipulation task at runtime. Based on the configuration, we compute the velocity of the end-effector.  The different ways to compute the goal configuration are:
\begin{itemize}
\item For the task of manipulating deformable objects to a single state $m^*$, the goal configuration can be represented simply by the visual feature of the desired state $s^*=s(I^*)$. 
\item For the task of manipulating deformable objects to a hidden state $h^*$, which can be represented by a set of states of the object $h^* = \{m_1, \cdots, m_n\}$
as a set of visual features $\{s(I_1),\cdots,s(I_n)\}$. We modify the formulation in  Equation \ref{eq:interaction} to compute $v$ as:
\begin{equation}
\label{eq:f_h1}
v = -\lambda \min_i (F(s(I) - s(I_j)))
\end{equation}
\item For a complex task, which can be represented using  a sequential set of states $\{m_1, \cdots, m_n\}$, we estimate the sequential cost of each state as $c(m_i)$. We use a modification that tends to compute the state with the lowest sequential cost:
\begin{equation}
\label{eq:f_h2}
i^* = \argmin_i (c(m_i) - \lambda F(s(I)-s(I_i))).
\end{equation}
After $i^*$ is computed,  the velocity for state $m_i$ is determined by $s(I_{i^*})$, and $m_i$ is removed from the set of goals for subsequent computations.
\end{itemize}

\subsection{Human Robot Interaction}
In many situations, the robot is working next to the human. The human is either grasping the deformable object or applying force. We 
classify the human-robot manipulation task using the hidden state goal $h^*$, where we need to estimate the human's action repeatedly. As the human intention is unknown to the robot, the resulting deformable material is assigned several goal states $\{m_1,\cdots,m_n\}$, which are determined conditionally by the action of human.   

\begin{figure}[t]
  \centering
  \includegraphics[width=0.4\textwidth]{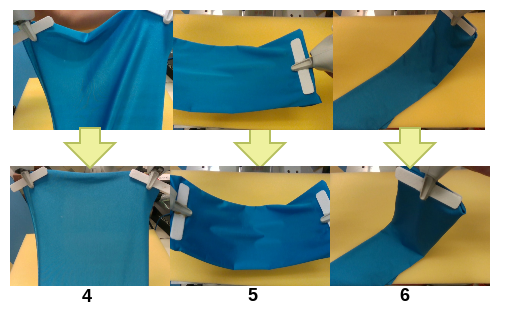}
  \caption{
  {\bf Manipulation Benchmarks:} We highlight three benchmarks corresponding to: (4) flattening; (5) placement; (6) folding.
  Top Row: The initial state of each task. Bottom Row: The goal state of each task. More details are given in Fig. 11.}
  \label{fig:benchmarks}
\end{figure}

\section{Implementation}
\label{sec:exp}
In this section,  we describe our implementation and the experimental setup, including the robot and the camera. We highlight the performance on several manipulation tasks performed by the robot only or robot-human collaboration. We also highlight the benefits of using HOW-features and the visual feedback dictionary.

\subsection{Robot Setup and Benchmarks}
\begin{figure}[t]
  \centering
  \includegraphics[width=0.35\textwidth]{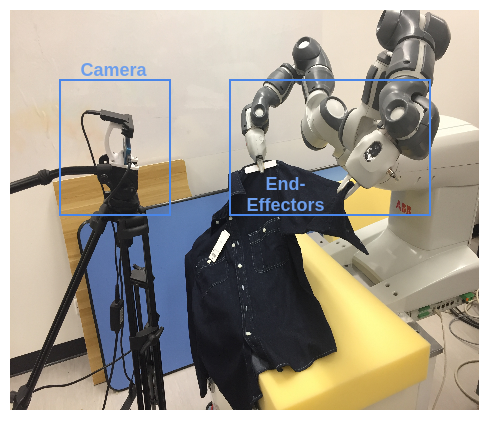}
  \caption{
  {\bf Setup for Manipulation Tasks:}
  {We use an $12$-DOF dual-arm ABB YuMi and an RGB camera to perform complex manipulation tasks using visual servoing, with and without humans.}
  }
  \label{fig:setup}
  \vspace*{-0.1in}
\end{figure}

Our algorithm was implemented on a PC and integrated with an 
 ABB YuMi dual-arm robot with 12-DOF to capture $\{r^{(t)}\}$  and perform manipulation tasks. We  use a RealSense camera is used to capture the RGB videos at  $(640\times480)$ resolution. In practice, we compute the Cartesian coordinates of the end-effectors of the ABB YuMi as the controlling configuration $r \in \mathbb{R}^6 $ and use an inverse kinematics-based motion planner~\cite{beeson2015trac} directly. The setup is shown in Figure \ref{fig:setup}.

\begin{table}[t]
\resizebox{0.48\textwidth}{!}{
\begin{tabular}
{llll }
    \hline
     Benchmark\# & Object  & Initial State & Task/Goal \\ \hline
     1 & towel & unfold in the air & fold with human\\ \hline
     2 & shirt & shape (set by human) & fixed shape \\ \hline
     3 & unstretchable cloth & position (set by human) & fixed shape \\ \hline
     4 & stretchable cloth & random  & flattening \\ \hline 
    5 & stretchable cloth & random   & placement \\ \hline
    6 & stretchable cloth & unfolded shape on desk & folded shape \\ \hline
  \end{tabular}
 }
\caption{
	{\bf Benchmark Tasks:}
	{We highlight various complex manipulation tasks performed using our algorithm. Three of them involve human-robot collaboration and we demonstrate that our method can handle external forces and perturbations applied to the cloth. We use cloth benchmarks of different material characteristics. The initial state is a random configuration or an unfolded cloth on a table, and we specify the final configuration for the task. The benchmark numbers correspond to the numbers shown in Fig.   \ref{fig:benchmark1} and Fig. \ref{fig:benchmarks}}     }
\label{tab:benchmarks}
\end{table}

To evaluate the effectiveness of our deformable manipulation framework, we use $6$ benchmarks with different clothes, which have different material characteristics and shapes.  Moreover, we use different initial and goal states depending on the task, e.g. stretching or folding. 
The details are listed in Table \ref{tab:benchmarks}.
In these tasks, we use three different forms of goal configurations for the deformable object manipulations, as discussed in Section \ref{sec:goal}.
 For benchmarks 4-6, the task corresponds to manipulating the cloth without human participation and we specify the goal configurations. For benchmarks 1-3, the task is to manipulate the cloth with human participation. The human is assisted with the task, but the task also introduces external perturbations.
Our approach makes no assumptions about the human motion, and only uses the visual feedback to guess the human's behavior.
In benchmark 1, the robot must anticipate the human's pace and forces for grasping to fold the towel in the air.
In benchmark 2, the robot needs to process a complex task with several goal configurations when performing a folding task.
In benchmark 3, the robot is asked to follow the human's actions to maintain the shape of the sheet. 
All 6 benchmarks are defined with goal states/features of the cloth, regardless of whether if there is a human moving the cloth or not. Because different states the cloth can be precisely represented and corresponding controlling parameters can be computed, the robot can perform complicated tasks as well.

\subsection{Benefits of HOW-feature}
There are many standard techniques to compute low-dimensional features of deformable models from RGB data known in computer vision and image processing. These include standard HOG and color histograms. We evaluated the performance of HOW-features along with the others and also explore the combination of these features.
The test involves measuring the success rate of the manipulator in moving towards the goal configuration based on the computed velocity, as given by Equation \ref{eq:velocity}.  
We obtain best results in our benchmarks using HOG+HOW features. The HOG features capture the edges in the image and the HOW captures the wrinkles and deformation, so their combination works well. 
For benchmarks 1 and 2, the shapes of the objects changes significantly and HOW can easily capture the deformation by enhancing the edges. 
For benchmarks 3, 4, and 5, HOW can capture the deformation by the shadowed area of wrinkles.
For benchmark 6, the total shadowed area continuously changes through the process,  in which the color histogram describes the feature slightly better.   
\subsection{Benefits of Sparse Representation}
The main parameter related to the visual feedback dictionary that governs the performance is its size. At runtime, it is also related to the choice of the slack variable in the sparse representation.
As the size of the visual feedback dictionary grows, the velocity error tends to reduce. However,  after reaching a certain size, the dictionary contributes less to the control policy mapping. That implies that there  is redundancy in the visual feedback dictionary. 

The performance of sparse representation computation at runtime is governed by the slack variable $\alpha$ in Equations \ref{alpha1} and \ref{alpha2}. This parameter provides a tradeoff between data fitting and sparse solution and governs the velocity error between the desired velocity $v^*$ and the actual velocity,
 $||v-v^*||_2$. In practice,  $\alpha$ affects the convergence speed. If $\alpha$ is small, the sparse computation has little or no impact and the solution tends to a common linear regression. If $\alpha$ is large, then we suffer from over-fitting.


\begin{figure}[t]
  \centering
  \includegraphics[width=0.4\textwidth]{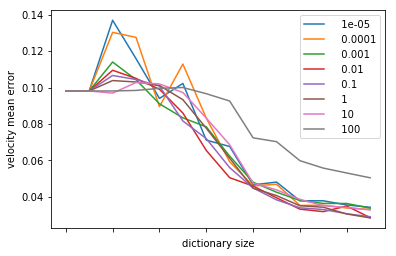}
  \caption{
  {\bf Parameter Selection for Visual Feedback Dictionary and Sparse Representation:} We vary the dictionary size on the X-axis and compute the velocity error for different values of $\alpha$ chosen for sparse representation for benchmark 4.
  }
  \label{fig:parameters}
  \vspace*{-0.1in}
\end{figure}

\begin{table}[t]
  \resizebox{0.5\textwidth}{!}{
  \begin{tabular}{lccc}
    \hline
    Feature & Benchmark 4 & Benchmark 5  & Benchmark 6  \\  \hline
    HOG & 71.44\%  & 67.31\%  & 82.62\%  \\ \hline 
    Color Histograms & 62.87\% &  53.67\%  & \bf 97.04\%    \\ \hline
    HOW & 92.21\%  &  71.97\% & 85.57\% \\ \hline
    HOW+HOG & \bf 94.53\%  & \bf 84.08\%  & 95.08\%  \\ \hline
  \end{tabular}
  }
\caption{{\bf  Comparison between Deformable Features:} 
We evaluated the success rate of the manipulator based on different features in terms of reaching the goal configuration based on the velocity computed using those features. For each experiment, the number of goals equals the number of frames. There are 393, 204 and 330 frames in benchmarks 4, 5, and 6, respectively. Overall, we obtain the best performance by using  HOG + HOW features. 
}
    \label{tab:comparison}
\end{table}















\section{Conclusion, Limitations and Future Work}
We present an approach to manipulate deformable materials using visual servoing and a precomputed feedback dictionary. We present a new algorithm to compute  HOW-features, which capture the shape variation and local features of the deformable material using limited computational resources. The visual feedback dictionary is precomputed using sampling and clustering techniques and used with sparse representation to compute the velocity of the controller to perform the task. We highlight the performance on a 12-DOF ABB dual arm and perform complex tasks related to stretching, folding, and placement. Furthermore, our approach can also be used for human-robot collaborative tasks. 

Our approach has some limitations. The effectiveness of the manipulation algorithm is governed by the training data of the specific task, and the goal state is defined by the demonstration. Because HOW-features are computed from 2D images, the accuracy of the computations can also vary based on the illumination and relative colors of the cloth. There are many avenues for future work. Besides overcoming these limitations, we would like to make our approach robust to the training data and the variation of the environment. Furthermore, we could use a more effective method for collecting the training data and generate a unified visual feedback dictionary for different tasks.
 
\section{Acknowledgement}
Jia Pan and Zhe Hu were supported by HKSAR General  Research  Fund (GRF) CityU 21203216, and NSFC/RGC Joint Research Scheme (CityU103/16-NSFC61631166002).


\bibliographystyle{IEEEtran}
\bibliography{ref}

\end{document}